\pdfoutput=1
\documentclass{article}
\usepackage{arxiv}

\usepackage[utf8]{inputenc} 
\usepackage{url}            
\usepackage{booktabs}       
\usepackage{amsfonts}       
\usepackage{nicefrac}       
\usepackage{microtype}      
\usepackage{graphicx}

\usepackage{url,hyperref,lineno,microtype,subcaption}
\usepackage{mathtools}
\usepackage[mathscr]{euscript}
\DeclareSymbolFont{rsfs}{U}{rsfs}{m}{n}
\DeclareSymbolFontAlphabet{\mathscrsfs}{rsfs}
\usepackage{amsfonts, amsmath, amsthm, amssymb}
\usepackage{soul}
\usepackage{physics}
\usepackage{caption}

\usepackage{times}
\usepackage{verbatim}
\usepackage{textcomp}
\usepackage{nameref}
\usepackage{xcolor}
\emergencystretch=\maxdimen
\usepackage{mathrsfs}

\usepackage{algorithm}
\usepackage{algorithmic}

\usepackage{float}
\usepackage{multicol}

\usepackage{hyperref} 

\title{Towards Understanding the Effect of Leak in Spiking Neural Networks}

\author{
  Sayeed Shafayet Chowdhury\thanks{equal contribution, authors listed in alphabetical order of surname.} \\
  Department of Electrical and Computer Engineering \\
  Purdue University\\
  West Lafayette, IN 47905 \\
  \texttt{chowdh23@purdue.edu} \\
  \And
 Chankyu Lee$^{*}$\\
  Department of Electrical and Computer Engineering \\
  Purdue University\\
  West Lafayette, IN 47905 \\
  \texttt{lee2216@purdue.edu} \\
   \And
 Kaushik Roy \\
  Department of Electrical and Computer Engineering \\
  Purdue University\\
  West Lafayette, IN 47905 \\
  \texttt{kaushik@purdue.edu} \\
}
   
\begin{document}

\maketitle

\begin{abstract}
Spiking Neural Networks (SNNs) are being explored to emulate the astounding capabilities of human brain that can learn and compute functions robustly and efficiently with noisy spiking activities. A variety of spiking neuron models have been proposed to resemble biological neuronal functionalities. With varying levels of bio-fidelity, these models often contain a leak path in their internal states, called membrane potentials. While the leaky models have been argued as more bio-plausible, a comparative analysis between models with and without leak from a purely computational point of view demands attention. In this paper, we investigate the questions regarding the justification of leak and the pros and cons of using leaky behavior. Our experimental results reveal that leaky neuron model provides improved robustness and better generalization compared to models with no leak. However, leak decreases the sparsity of computation contrary to the common notion. Through a frequency domain analysis, we demonstrate the effect of leak in eliminating the high-frequency components from the input, thus enabling SNNs to be more robust against noisy spike-inputs.
\end{abstract}

\section{Introduction}
Over the past few years, the advancements of deep artificial neural networks (ANNs) have led to remarkable success in various cognitive tasks ($e.g.,$ vision, language and behavior). In some cases, neural networks have outperformed the conventional algorithms and achieved human-level performance \cite{simonyan2014very, Silver2016}. However, recent ANNs are becoming extremely compute-intensive and often do not generalize well to previously unseen data during training. On the other hand, human brain can reliably learn and compute intricate cognitive tasks with only a few watts of power budget. Recently, Spiking Neural Networks (SNNs) have been explored toward realizing robust and energy-efficient machine intelligence guided by the cues from neuroscience experiments \cite{roy2019towards}. 

SNNs are categorized as the new generation neural networks \cite{maass1997networks} based on their neuronal functionalities. A variety of spiking neuron models largely resemble biological neuronal mechanisms, which transmit information through discrete spatio-temporal events (or spikes). These spiking neuron models can be characterized by their internal state called the membrane potential. A spiking neuron integrates the inputs over time and fires a spike-output whenever the membrane potential exceeds a threshold. Computational models for spiking neurons use a Leaky Integrate and Fire (LIF) model, which has a built-in leaky behavior in the membrane potential, or use simpler Integrate and Fire (IF) with no leak in the membrane potential \cite{Dayan}. Since biological neuron models have been reported to contain leak in the membrane potential \cite{snutch2007sodium}, it would be important to quantitatively analyze the advantages and disadvantages of using leaky behavior.

To that end, we focus on two aspects of the leak effect on SNN models: robustness and spiking sparsity. Ideally, the neural network models are expected to predict reliable outcomes for unseen or even noisy data under sparse spiking events. In addition, compared to ANNs, the main advantage of SNNs is the energy-efficient event-based computing capability, in which the synaptic operations occur only when spike-inputs arrive. To that effect, the computational efficiency of SNNs considerably improves as spike signals become sparser for specialized SNN hardware platforms such as TrueNorth \cite{truenorth} and Loihi \cite{davies2018loihi}. 

Although various models have been proposed that resemble realistic biological neuronal mechanisms \cite{hodgkin1952currents, fitzhugh1961impulses, brette2005adaptive}, they are often too complex from a computational point of view. Also, there is a lack of understanding as to how each of the factors determining the biological neuronal response can be effectively used in learning. Hence, we investigate the general and simple IF and LIF neuron models \cite{Dayan} that are analytically tractable. We present a comprehensive and comparative analysis between models with and without leak to delve deeper into the role that leak plays in learning. The main contributions of this work are as follows, 

\begin{itemize}
\setlength{\itemindent}{0em}

\item A theoretical analysis of the first-order phenomenological LIF neuron model is introduced to investigate its low-pass filtering effect. As a step toward this goal, from frequency domain analyses, we show that the presence of leak helps to cut-off some of the input components beyond a certain frequency, thereby aiding the networks to predict more robust outcomes for noisy spike-inputs.

\item We examine the effect of leak on compute requirements in multi-layered SNNs. Compared to SNNs with IF model, the ones with LIF model converges with decreased sparsity of spike signals when trained with surrogate-gradient based backpropagation, resulting in reduced computational efficiency. Notably, leak is only varied from one model to the other, whereas, during training any particular model, leak is kept fixed. 

\item We conduct experiments to validate the robustness of multi-layered SNNs with IF and LIF neuron models using popular vision datasets including SVHN and CIFAR-10. Furthermore, we analyze the improved performance of LIF models by investigating the frequency spectrum of spikes and how well the network generalizes to previously unseen data.

\end{itemize}

\begin{figure}[b]
\centering
\includegraphics[width=\columnwidth]{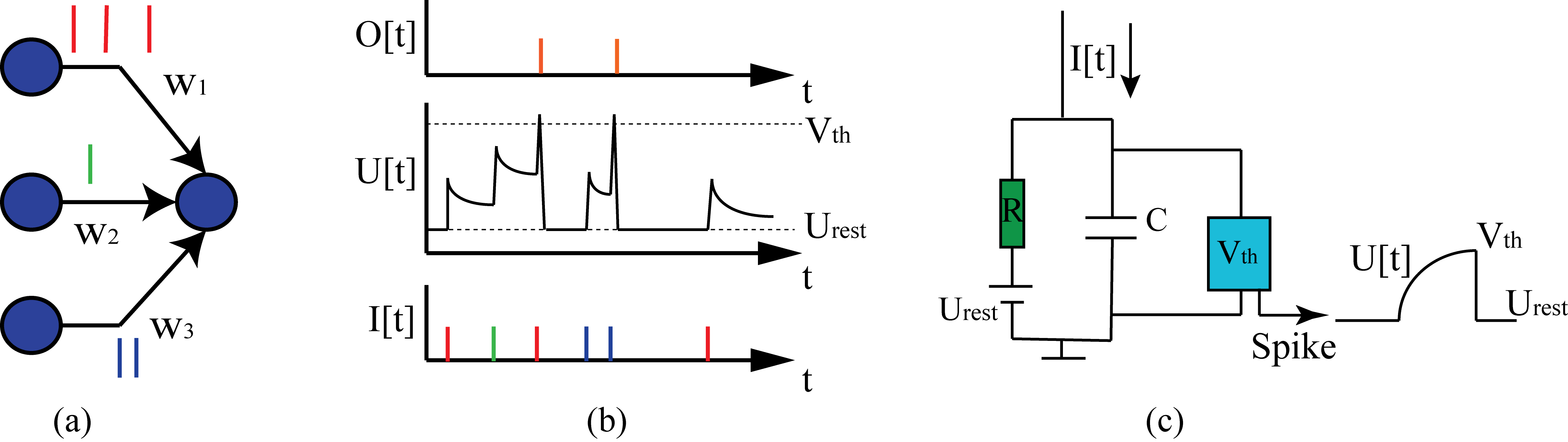}
\caption{An LIF neuron, (a) a schematic connection between three pre-neurons to one post-neuron, (b) temporal dynamics of membrane potential in the post-neuron, (c) equivalent circuit model of the LIF neuron.}
\label{fig:lif}
\vspace{-5mm}
\end{figure}

\section{Spiking Neural Network Fundamentals}
\subsection{Spiking Neuron Model}
The spiking neurons (generally modeled as IF or LIF) are fundamental units in SNNs. The sub-threshold dynamics of an LIF neuron is governed by,
\begin{equation}\label{eqn:1}
\tau_{m} \frac{dU}{dt}= -(U-U_{rest}) + RI,~~~~U\leq V_{th}
\end{equation}
where $U$ is the membrane potential, $I$ denotes the input current that represents the weighted summation of spike-inputs, $\tau_{m}$ indicates the time constant for membrane potential decay, $R$ represents membrane leakage path resistance and $U_{rest}$ is resting potential. Fig.~\ref{fig:lif} depicts the dynamics of LIF neuron and an equivalent circuit model. The input current is accumulated in the membrane potential that decays exponentially over time. The degree of exponential decay is determined by the membrane time constant ($\tau_{m}$). When the membrane potential exceeds the firing threshold ($V_{th}$), the neuron is triggered to emit an output-spike and resets the membrane potential to the resting state. The spike-output can be represented as, 
\begin{equation*}
O[t]=\left\{
                \begin{array}{ll}
                  1,~~~\text{$if~U[t]>V_{th}$}\\
                  0,~~~\text{$otherwise$}
                \end{array}
              \right.
              \tag{2}
\end{equation*}
where $O[t]$ and $U[t]$ denote the spike-output and the membrane potential, respectively, at time instant $t$. The neuronal dynamics in Eqn.\ref{eqn:1} can be represented by an equivalent RC circuit model \cite{epfl} as illustrated in Fig.~\ref{fig:lif}(c). The parallel RC branch acts as a low-pass filter \cite{fourcaud2003spike}, which has the membrane time constant ($\tau_{m}$) as $RC$ where $C$ is the membrane capacitance. 

\subsection{Frequency Domain Analyses}
In this subsection, the response of an LIF neuron model is analyzed in relation to the membrane time constant ($\tau_m$). We investigate the role of leaks in filtering out some of the signal components in the high frequency range when driven by white Gaussian noise. In order to quantify the low-pass filtering effect, we employ the coherence function, $C(\omega)$ which is a commonly used metric in signal processing \cite{proakis2001digital} to estimate the power transfer from the input to the output. When the input to a system is $s(t)$ and the corresponding output is $x(t)$, the coherence between them is defined as,
\begin{equation*}
C_{x,s}(\omega)=\frac{|S_{x,s}(\omega)|^2}{S_{x,x}(\omega)S_{s,s}(\omega)},
\tag{3}
\end{equation*}
where $S_{x,s}(\omega)$ is the cross-spectrum of output ($x$) with input ($s$), $S_{x,x}(\omega)$ and $S_{s,s}(\omega)$ are the autopower spectrum of $x(t)$ and $s(t)$, respectively. To study the response of the neuron model described by Eqn.~\ref{eqn:1}, we measure the coherence as a function of frequency. 
We model the inputs to the neuron as white Gaussian noise current and derive the corresponding coherence between the noise input and the output spike train. The resulting coherence function $C_{x,s}(\omega)$ is as follows:
\begin{equation*}
\small{
C_{x,s}(\omega)=\frac{2D_{st}}{{D}}\frac{r_0\omega ^2}{1+\omega^2}\frac{\left |\mathscrsfs{D}_{i\omega-1} \left(\frac{\mu-V_{th}}{\sqrt{D}}  \right)-e^\Delta \mathscrsfs{D}_{i\omega-1} \left(\frac{\mu-U_{rest}}{\sqrt{D}}  \right)\right |^2}{\left |\mathscrsfs{D}_{i\omega} \left(\frac{\mu-V_{th}}{\sqrt{D}}  \right)\right|^2-e^{2\Delta} \left |\mathscrsfs{D}_{i\omega} \left(\frac{\mu-U_{rest}}{\sqrt{D}}  \right) \right |^2}},
\tag{4}
\end{equation*}
where $D_{st}$ is the intensity of the white noise stimulus, $D$ is total noise intensity (for our case $D$=$D_{st}$), $r_0$ is the output firing rate, $\mathscrsfs{D}(x)$ is a parabolic cylinder function, $\mu$ is a parameter denoting DC part of the input (defined in supplementary section) and $\Delta= \frac{U_{rest}^2-V_{th}^2+2\mu (V_{th}-U_{rest})}{4D}$. The detailed derivations of Eqn.~1-Eqn.~4 
are provided in the supplementary information.

\begin{figure}
\centering
\includegraphics[width=.3\columnwidth]{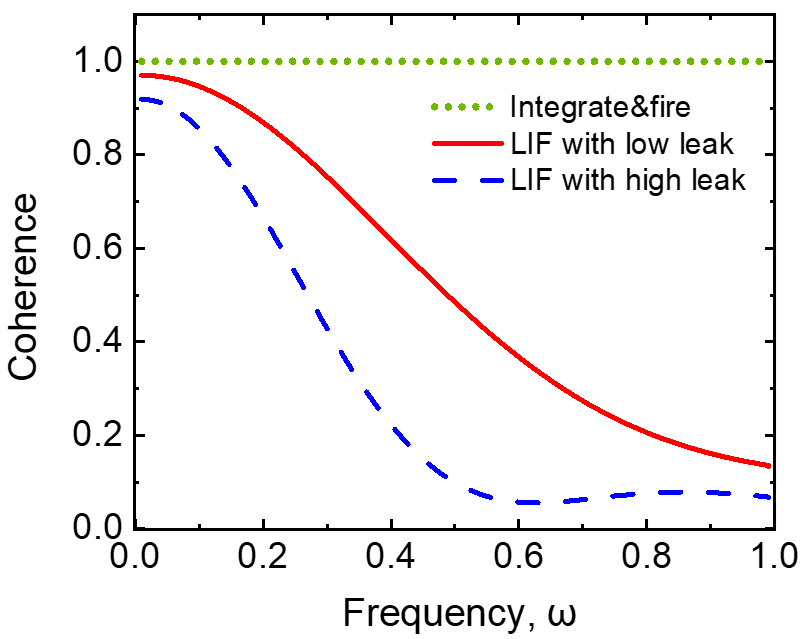}
\caption{Illustration of frequency response for IF and LIF neuron models. The horizontal and vertical axes represent the frequency components and coherence function, respectively.}
  \label{fig:coherence}
  
\end{figure}

To analyze the frequency responses of the neuron model, the coherence functions of the IF and the LIF models with high and low leak cases in relation to frequency ($\omega$) are plotted in Fig.~\ref{fig:coherence}. This figure shows that IF model (green) transmits all input components to the outputs across the entire frequency spectrum. On the other hand, for LIF models (red and blue), the coherence function decreases as the frequency increases, thereby cutting-off the high frequency components propagating to the output. Hence, contrary to the IF model, the LIF model can negate the noise input components beyond a certain frequency limit. Similar low-pass filtering of information for LIF neurons has been reported in \cite{fourcaud2003spike, sharafi2013information}. The authors in \cite{connelly2016thalamus} also discussed similar characteristics of LIF neurons from a neuroscience perspective. Drawing inspirations from such phenomenon, our next goal is to explore whether the low-pass filtering effect can enable multi-layered SNNs with leaky neuron models to be more robust against noisy inputs. The following subsections focus first on the training methodology adopted in this work, followed by the noisy input generation methods and corresponding experiments.

\subsection{Gradient Descent Learning in SNNs}
The gradient-based method, namely backpropagation (BP) learning \cite{rumelhart1985learning}, is widely employed for training traditional deep ANNs. While ANN neuron models with continuous functions (such as $sigmoid$, $tanh$ or $ReLU$) are compatible with the gradient-based learning, it has been a challenge to directly train SNNs with BP method in their native form. This is due to the spike-output being binary-valued ($i.e.$, zero or one), which renders the spike generation function non-differentiable and discontinuous. To get around this issue, standard BP has been adapted for the spike-based learning domain which we refer to as `spike-based backpropagation'. The spike-based BP method overcomes the discontinuous spiking functionality by approximately estimating the surrogate gradient of spike generation function. Several surrogate gradient methods have been introduced in the literature \cite{lee2016training,mostafa2017supervised,huh2018gradient}. In this work, we employ the LIF neuronal surrogate gradient function that accounts for the leaky behavior as proposed in \cite{lee2019enabling}. 

The training procedure is composed of two phases ($e.g.,$ forward and backward). In the forward phase, the hidden layer neurons accumulate the weighted sum of spike-inputs in the membrane potential. When this potential exceeds the threshold, the neuron fires a spike-output and resets the potential to the resting state (zero). Otherwise, membrane potential decays exponentially. The final layer neurons do not generate spike output and decay over time, accumulating a weighted sum of spike-inputs. At the last time step, the final prediction outcomes are estimated by dividing the final layer membrane potential ($U_L[T]$) by the total number of time-steps ($T$). Then, the final errors are evaluated by comparing the final prediction outcomes with the ground truth ($label$). The loss function ($Loss$) is obtained by computing the summation of squared error as shown below, 

\begin{equation*}
Loss = \frac{1}{2} (\frac{U_L[T]}{T} - label)^2, ~~~
\frac{\partial O_l[t]}{\partial U_l[t]}= \frac{1}{V_{th}+\epsilon}(O_l[t]>0),
\tag{5}
\end{equation*}

      \begin{algorithm}[H]
        \caption{Poisson spike generation scheme under noise}
        \label{alg:Algo1}
        \begin{algorithmic}
        \STATE {\bfseries Input:} pixel-based inputs ($inputs$), total number of time steps ($\#timesteps$), external random noise ($\xi$), uniform random number ($\mathscr{X}$)
        \STATE {\bfseries Output:} spike-based inputs ($O_1[t]$)
        \FOR{$t \leftarrow 1$ {\bfseries to} $\#timesteps$}
            \IF{$Scenario~1$}
            \STATE //~External noise ($\xi$) is added to input pixel
            \STATE $inputs_c$ = $inputs+\xi$
            \STATE //~If noisy input ($inputs_c$) is greater than uniform random number, a spike-input ($O_1[t]$) is generated
            \IF{$inputs_c > \mathscr{X}$}
            \STATE $O_1[t] = 1$
            \ELSE
            \STATE $O_1[t] = 0$
            \ENDIF
            \ELSIF{$Scenario~2$}
            \STATE //~External noise ($\xi$) is added to input channel
            \IF{$inputs > \mathscr{X}$}
            \STATE $O_1[t] = 1 + \xi$
            \ELSE
            \STATE $O_1[t] = \xi$
            \ENDIF
            \ENDIF
        \ENDFOR
        \end{algorithmic}
      \end{algorithm}
  
where $\frac{U_L[T]}{T}$ is the final prediction outcome. In the backward phase, the final errors are propagated backward while unrolling the network in time using the surrogate gradient method. This procedure is often regarded as Backpropagation Through Time (BPTT) \cite{werbos1990backpropagation}. The surrogate gradient of LIF neuronal function is computed by combining the straight through estimation \cite{bengio2013estimating} and leak correctional term ($\epsilon$) as given by the second equation in Eqn.~5. Here, the straight-through estimation ($i.e., \frac{1}{V_{th}}$) calculates the derivative of IF neuronal function and $\epsilon$ compensates the leaky effect of the membrane potential. 
Finally, the network parameters are updated based on the partial derivatives of the loss with respect to weights for all discrete time steps. The trained SNNs can incorporate temporal and leak statistics from direct spike-inputs over time. The pseudo code of the spike-based BP learning can be found in the supplementary material.

\section{Poisson Spike Generation under Noisy Environments}
In section 4.2, the spike-inputs with external random noise are used for experimentally evaluating the noise robustness ($i.e.,$~the capability of maintaining a certain prediction accuracy under stochastic perturbations) of multi-layered SNNs. Keeping that goal in mind, here we explain the noisy spike-input generation methods used in our work. Specifically, two different sources of random noise are considered, namely $Gaussian~noise$ and $Impulse~noise$ \cite{hendrycks2019benchmarking}. Under each noise source, two noise injection scenarios are introduced for producing the noisy spike-inputs. Each noisy spike-input generation procedure is depicted in Algorithm \ref{alg:Algo1}.

For scenario 1, an independent random noise is added to a image pixels at each time step. The combination of pixel input and noise is compared with an uniformly distributed random number to generate Poisson-distributed spike-inputs. Hence, for a given period of time, the stream of spike-inputs incorporates the noise over time. For scenario 2, an independent random noise is added (at each time step) to the Poisson spike trains generated from the original image pixels. The major difference between two scenarios is whether the random noise is added before or after comparing with a random number (Poisson spike generation process). Note, in scenario 1, spikes are generated as a post-process of adding noise to image pixels, making the input spike train strictly binary-valued, but in scenario 2, noise is added directly to the spikes, so
the resultant noisy spikes contain perturbations around their clean spike values (0 or 1). The random noise injection process is performed in the input layer only.

\vspace{-2mm}

\section{Experiments}

\subsection{Experimental Setup}
We examine the robustness of multi-layered SNNs against noisy spike-inputs on two standard vision benchmarks, namely SVHN and CIFAR-10, which are composed of color (three-dimensional) inputs. We experiment with multi-layered SNN models, which comprise of 32$\times$32 color inputs, convolutional (C) layers with 3$\times$3 weight kernels, average-pooling (P) layers with fixed 2$\times$2 kernel followed by fully-connected (FC) layers. The details of the chosen SNN models are as follows: model used for CIFAR-10 is (32$\times$32-64C3-64C3-2P-128C3-128C3-2P-256C3-256C3-256C3-2s-1024FC-10o) and model used for SVHN is (32$\times$32-64C3-64C3-2P-256C3-256C3-256C3-2s-1024FC-10o). We follow the training protocols as described in \cite{lee2019enabling}. Each network model with different membrane time constant is independently trained with clean training data. Note, the membrane time constant is not considered as a trainable parameter and remains fixed during training and testing. All network models are trained with mini-batch spike-based BP for 150 epochs with a batch size of 64, while decreasing the learning rate at $70^{th}$ and $100^{th}$ epoch. After normalizing each image sample to zero mean and scaling to the range [-1, 1], Poisson spike trains are generated for 100 time-steps during training and testing. The reported results are the average score from three independently trained networks. We implemented the multi-layered SNNs using Pytorch deep learning package.

\begin{figure*}[h]
\vspace{-1mm}
\centering
\includegraphics[width=\textwidth]{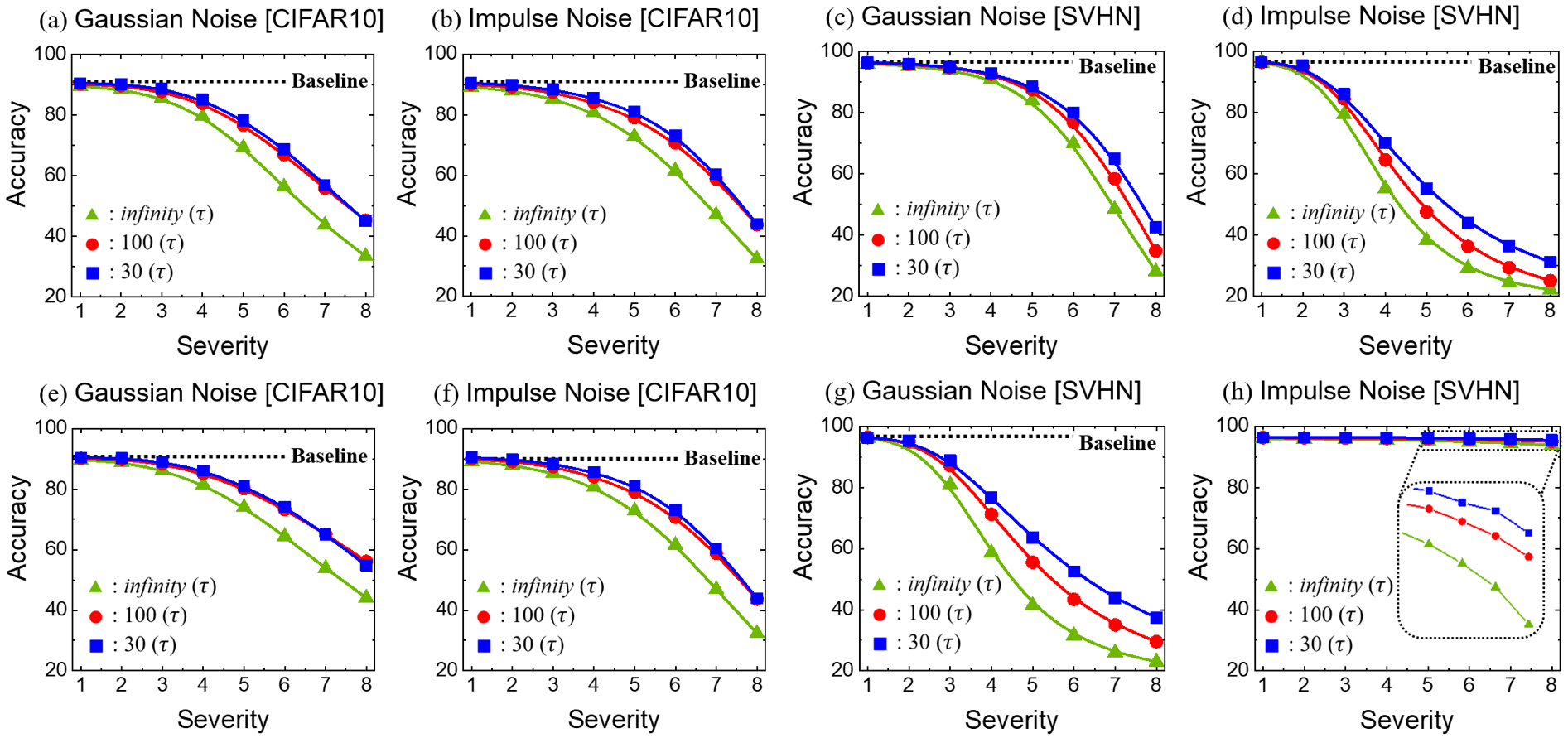}
\caption{Classification accuracy at each level of noise severity. The horizontal and vertical axes present the input noise severity and classification accuracy, respectively. (a,b,c,d) Results from noisy input generation scenario 1. (e,f,g,h) Results from noisy input generation scenario 2.}
\label{fig:ttacc_lee}
\vspace{-3mm}
\end{figure*}

\subsection{Robustness against Noisy Spike-inputs}
First, we compare the noise robustness results with different membrane time constants ($e.g, \tau_m$= $30, 100$ and $infinity$). The LIF neuron models are associated with relatively smaller membrane time constants ($e.g., \tau_m$= $30$ and $100$) compared to IF neuron model with an infinitely large membrane time constant ($e.g., \tau_m$= $infinity$). The robustness of each SNN model is measured in terms of the stability of the classification accuracy against noisy spike-inputs. The performance of SNNs are scored across eight severity levels with each noise type ($e.g.,$ $Gaussian~noise$ and $Impulse~noise$). The severity level indicates the strength of input noise.

\begin{table}[t]
\caption{Comparison between the network models with different leak amounts. The first row corresponds to baseline accuracy. The second and the third rows correspond to the sum-squared errors averaged over 130–150 epochs for testing and training data, respectively. The fourth and the fifth rows correspond to average spiking activity and the total number of synaptic operations, respectively.}
\begin{center}
\label{table:sse}
\begin{tabular}{lcccccccr}
\toprule
\textbf{Dataset}      & \multicolumn{3}{c}{\textbf{CIFAR-10}} & \multicolumn{3}{c}{\textbf{SVHN}} \\ 
$\tau_m$ & 30       & 100       & inf       & 30      & 100      & inf     \\ 
\midrule
$Accuracy~(\%)$      &  89.65    &   90.19    &    90.3    &    96.12     &  96.32    &  96.32        \\
$SSE_{Test}$      &  \textbf{2.93}    &   3.45    &     3.83    &    \textbf{0.72}     &  0.75      &  0.82        \\
$SSE_{Train}$     &  \textbf{1.88}   &    1.92   &  2.2         &   \textbf{1.26}   &  1.4    & 1.6         \\ 
$Spikes~(\%)$      &  \textbf{9.45}    &   5.26    &    4.94    &    \textbf{14.07}     &  12.09      &  11.85        \\
$\# Synaptic~Ops$      &  \textbf{1.59E9}    &   7.92E8    &    7.18E8    &    \textbf{3.99E9}     &  3.88E9      &  3.79E9        \\

\bottomrule
\end{tabular}
\end{center}
\vspace{-5mm}
\end{table}

In both benchmark datasets (CIFAR-10, SVHN), the baseline testing accuracy is almost the same under different leaks as presented in the first row of Table \ref{table:sse}. Fig.~\ref{fig:ttacc_lee} shows the accuracy results with increasing level of noise severity across different benchmarks (first row: noisy spike generation scenario 1, second row: noisy spike generation scenario 2). For both the noisy spike generation scenarios, SNNs with LIF neurons (blue, red) achieve improved noise robustness whereas the ones with IF neurons (green) suffer from severe accuracy degradation for high noise severity levels as displayed in Fig.~\ref{fig:ttacc_lee}. We would like to mention here that all network models are trained on clean spike-inputs, but tested with noisy ones. The models trained with the highest amount of leak (blue) retain the baseline accuracy to a greater extent compared to a non-leaky model. The LIF model with $\tau_m$= $100$ shows relatively higher accuracy degradation compared to one with $\tau_m$= $30$. However, both models show improved robustness compared to the IF model. These trends hold true for all noisy spike-input generation scenarios. In our experiments, we observed that training loss diverges when the chosen membrane time constant is too small ($\tau_m<30$). In this case, the spiking activities decrease severely due to extremely high leak while passing through the layers, causing convergence issues in multi-layered SNN training \cite{lee2019enabling}. 
However, for converged network models, SNNs with leaky neurons exhibit better stability against noisy spike-inputs.

\begin{figure}[h]
\centering
\includegraphics[width=0.5\columnwidth]{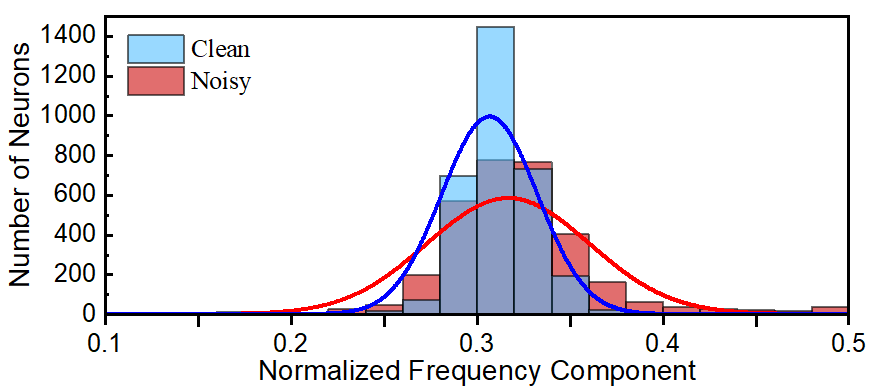}
\caption{Histogram of the spectrum of spike trains per image for clean and noisy (Gaussian noise) inputs with corresponding distribution curves.}
\label{fig:ip_spectr}
\end{figure}


\subsection{Spectrum Analysis}
To analyze the improved noise robustness of LIF models, we perform a spectrum analysis of inputs and the corresponding network outputs for both clean and noisy data.
In general, the noise spectrum contains components over a wide frequency band. The single-sided spectra of input-spike trains (averaged over test samples) for the clean and the noisy cases are shown in Fig.~\ref{fig:ip_spectr}. It can be observed that the mean spectrum distribution with noisy spike-inputs remains roughly the same compared to the clean case. However, this spectrum of noisy data spreads over a wider band, resulting in more components in higher frequency bands. These changes in the spectrum distribution can significantly alter the spike patterns propagating through the layers compared to the clean input. As previously explained in section 2.2, the leaky neuron models only pass inputs with low frequency components. Hence, the leaky neuron models can eliminate some of the high-frequency noise components, thus helping to maintain the baseline performance. However, the low-frequency noise components pass through the LIF and IF models in a similar way. Thus, the accuracy degradation due to such components remains alike for both leaky and non-leaky neurons.

Next, let us consider the spectrum of the target output neuron (node corresponding to the ground truth label) in the final layer, since the changes concerning this output neuron largely determine the correct or wrong classification. For each image, we measure the average spectrum of the target output neuron and calculate the critical frequency up to which the significant power ($70\%$) of the total spectrum resides.
This critical frequency distribution is examined over all the samples and plotted in Fig.~5(a,b,c). Interestingly, with increasing amount of leak, we found that the mean spectrum shifts towards the left. We anticipate this shift towards the lower frequency band is owing to the inherent low-pass filtering effect of leak. 
The normalized mean critical frequency components for the target neuron corresponding to $\tau_{m}$= $infinity, 100$ and $30$ are 0.345, 0.317 and 0.255 respectively, for the clean testing samples, while for the noisy inputs (for noise severity level of five), the same frequency components become 0.35, 0.33 and 0.298, respectively. These outcomes along with Fig.~5(a,b,c) clearly indicate that frequency components of target neuron's output response become higher with noisy spike-inputs compared to the clean input case. As IF neurons have much wider pass-band, the higher frequency components are not filtered out as shown in Fig.~5(a), thus making the network more prone to have noise errors. In contrast, for the LIF models, most of the high frequency components are eliminated through the low-pass filtering effect as demonstrated in Fig.~5(b,c) which results in maintaining the baseline accuracy.



\begin{figure*}[h]
\centering
\includegraphics[width=0.93\columnwidth]{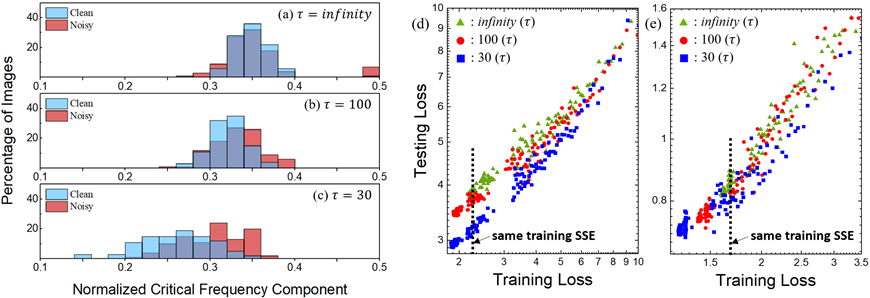}
\caption{Histogram of average normalized critical frequency components of target output neuron for (a) $\tau_{m}=infinity$, (b) $\tau_{m}=100$ and (c) $\tau_{m}=30$. The sum-squared errors of test data with respect to the ones of training data on (d) CIFAR-10 and (e) SVHN benchmarks. The horizontal and vertical axes present the sum-squared error (in log scale) on train and test data, respectively.}
\label{fig4}
\vspace{-3mm}
\end{figure*}

\subsection{Analyses of Generalization}
In order to ascertain the improved noise robustness from another perspective, we extend our analysis to generalization. We hypothesize that leaky neuron models enable SNNs to better generalize to previously unseen examples, and examine the impact of leak on generalization. While training multi-layered SNNs over 150 epochs, we recorded the sum of squared error (SSE) on the testing and the training samples to highlight the performance differences. As training progresses towards the end, we found that SNNs with LIF models yield lower testing SSE with the same training effort and reach lower final testing and training SSE than the ones with IF models. The second and the third rows of Table \ref{table:sse} present the testing and training SSE averaged over 130–150 epochs. We also analyze the testing SSE attained as a function of the training SSE. Fig.~5(d,e) shows the testing SSE with respect to the training SSE for different membrane time constants. We found that at the same training SSE, LIF models (blue, red) yield lower testing SSE than IF models (green), hinting towards better generalization. Notably, the advantage of better generalization is the mitigation of overfitting in large neural networks \cite{erhan2010does, lee2018training}.

\subsection{Input Activity Analysis}
While the enhanced robustness achieved through leaky neuron models is advantageous, it is also pivotal to consider the associated computations and energy costs of using LIF and IF models. To infer an output class, SNNs need the spike-inputs to be fed over a number of times steps, performing event-based synaptic operations that take place only when spike-inputs arrive. In this respect, the total number of synaptic operations is typically considered as a metric for benchmarking the computational costs in neuromorphic hardware \cite{davies2018loihi,merolla2014million}. This subsection explores the impact of leaky neuron models on the spiking sparsity and the number of computations, two critical factors that directly determine the computational efficiency of SNNs. 
In Table \ref{table:sse}, the fourth and fifth rows present the average spike activities and the total synaptic operations, respectively, for different leak parameters. We found that the overall spiking activities increase with a higher leak, thereby resulting in more synaptic computations. An important insight from here is that, with respect to the degree of leak, there exists a trade-off between noise robustness and compute requirements.

\begin{figure}[h]
\centering
\includegraphics[width=0.93\columnwidth]{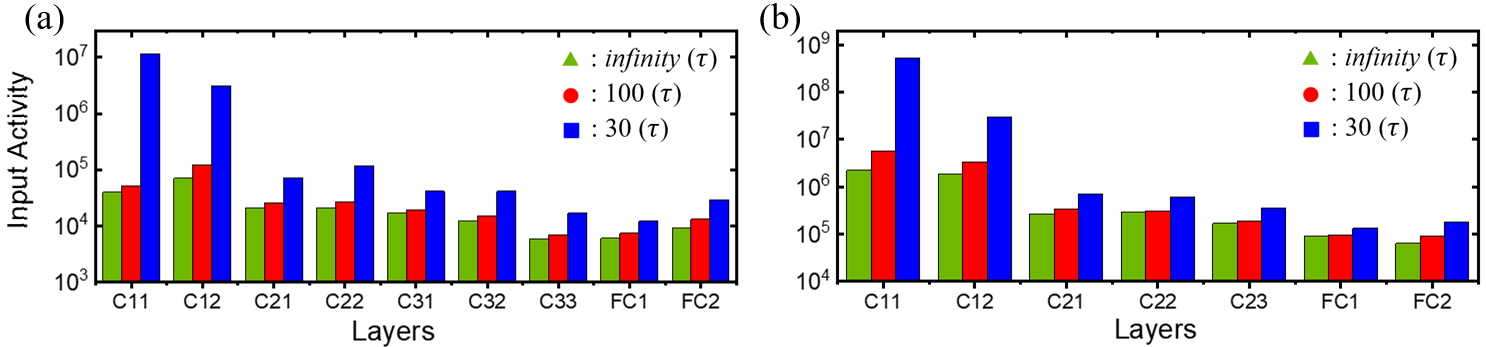}
\caption{Layer-wise Euclidean norm of the weighted sum of spike-inputs of multi-layered SNNs for (a) CIFAR-10 and (b) SVHN datasets.}
\label{fig:bar}
\vspace{-4mm}
\end{figure}

To investigate the reason behind the increased spiking activities with higher leak, we measure the Euclidean norm of the weighted sum of spike-inputs 
(referred to as `ENWSI' subsequently) over time for each hidden layer. We would like to note that ENWSI is representative of a combination of spiking activities and weights that determine the net input information to the corresponding layers. Fig.~\ref{fig:bar} illustrates the ENWSI for different membrane time constants ($e.g., \tau_m$= $30, 100$ and $infinity$). We found that SNNs with LIF neurons (blue, red) receive higher ENWSI across the layers than those with IF neurons (green). The model with the largest leak (blue) receives the highest ENWSI compared to the other models under consideration as evidenced in Fig. \ref{fig:bar}.

It is widely understood that if IF and LIF neurons were to receive the same weighted sum of inputs, the LIF neurons would produce comparatively lesser outputs due to their inherent leak. However, that would lead the LIF models not to have enough spikes in the deeper layers due to the layerwise gradual reduction in spiking activities. Hence, the resultant network would fail to converge with acceptable accuracy. To overcome the leak effect and to have sufficient spiking activities for proper training, spike-based BP training tailors the LIF models to increase the weighted sum of hidden layer input activities beyond what is needed for IF models. Consequently, LIF models converge to configurations with increased spiking activities, allowing for sufficient weighted sum of input activities in the deeper layers.

\vspace{-2mm}
\section{Related Works and Discussion}
\vspace{-2mm}
In the neuroscience literature, the existence of leak in biological neurons has been reported in the context of sodium ion channels \cite{snutch2007sodium, ren2011sodium}, synaptic transmission in visual cortex \cite{artun1998effect, millman2010self}, etc. 
SNN models take the bio-plausibility of leak into account through leaky neuron models \cite{hodgkin1952currents, fitzhugh1961impulses, brette2005adaptive}. In those models, leak acts as a hyperparameter that controls the decay of membrane potentials in the neurons over time.
However, the effect of leak in learning and the resultant neuronal responses have not been studied comprehensively, to the best of our knowledge. 
Recognizing this gap, in this paper we investigate IF and LIF neuron models to analyze the role that leak plays in learning and their impact on robustness and spiking sparsity. 

First, noise robustness has been a significant concern for neural networks, motivating a number of recent works. 
Data augmentation \cite{hendrycks2019benchmarking,hunsberger2015spiking,vasiljevic2016examining} and quantization \cite{han2015deep, panda2019discretization} have been shown to achieve robust performance in both SNNs and ANNs. However, data augmentation-based techniques usually do not generalize well to other types of noise than those used during training, necessitating expensive iterative training efforts using diverse augmented samples. Moreover, the input and weight quantization techniques are reported to be susceptible to error amplification due to enlarged quantization noise in multi-layered networks \cite{lin2019defensive}, leading to considerable loss of accuracy. Our work is pertinent in this respect, since the experimental results show that the leaky neuron models enable improved robustness against random noise, without the need for costly re-training procedures or error amplification. We attribute this enhanced robustness to the better generalization and low-pass filtering effects of the LIF neuron models. 

However, introducing leaks in the SNN models (while training with backpropagation) comes at the expense of higher spiking activities compared to the IF models. To that effect, with respect to the usage of leaky neuron models, there is a trade-off between noise robustness and computational efficiency. At $\tau_m$= $100$, SNNs with LIF neurons achieve substantially improved robustness compared to the ones with IF neurons while maintaining reasonable spiking sparsity. Training with higher leak ($\tau_m$= $30$) further improves the robustness; however, the spiking activities also increase considerably. To conclude, understanding of leak provides another knob for designing SNNs, enabling us to obtain a robustly trained network without sacrificing compute-efficiency significantly.

\vspace{-2mm}

\section{Broader Impact}
The scope of this work will be of interest to both the neuroscience and machine learning communities. From the neuroscience perspective, we want to comprehend the brain functionalities and the associated responses. On the other hand, from the machine learning point of view, we aim to implement brain-inspired  networks which are reliable and energy-efficient. To that end, we believe the findings of our work contribute a small step towards bridging the two seemingly disparate fields of neuroscience and machine learning. Our work will hopefully inspire the neuroscientists to effectively merge their knowledge with that of traditional machine learning practitioners in order to achieve a holistic understanding of the underlying models. This would enable more efficient and reliable neural computing for practical edge applications.

As we analyze the impacts of leak on robustness and sparsity, our study will be particularly useful for designing resource-constrained edge  applications in noisy environments ($e.g.,$ self-driving vehicles in adverse weather and rescue robots in disasters etc). Considering that leak is an essential bio-plausible element in SNN models, we believe a better understanding of its effects will help to design improved bio-inspired architectures by making optimal choices concerning the involved trade-offs. Additionally, there are opportunities to further explore the effect of leak, especially in terms of other more complicated neuron models ($e.g$., Hodgkin-Huxley \cite{hodgkin1952currents} and exponential integrate-and-fire \cite{brette2005adaptive}). Furthermore, an efficient algorithm-hardware co-design considering the leak impacts would  be of interest for future research directions.


\section*{Supplementary Material}

\section{Detailed Formulation of Coherence Function}
In this section, we present the derivations of the coherence function, $C(\omega)$ between the input stimulus for the Leaky Integrate and Fire (LIF) neuron and output spike train, 
guided by 
\cite{lindner2014low} and \cite{vilela2009input}. The discussion is divided into two parts: first we formulate the equation describing the neuron model in subsection 7.1, next using this formulated model, we derive equations to calculate coherence in subsection 7.2.
\subsection{LIF Model Equation}
 The dynamics of an LIF neuron is modeled as follows:
\begin{equation}\label{eqn:1}
\tau_{m} \frac{dU}{dt}= -(U-U_{rest}) + RI,~~~~U\leq V_{th}
\tag{1}
\end{equation}
where $U$ is the membrane potential, $I$ denotes the input current, $\tau_{m}$ indicates the membrane time constant, $R$ represents membrane resistance and $U_{rest}$ is the resting potential. Note, an equivalent parallel resistor-capacitor (RC) circuit model of the LIF neuron is illustrated in Fig. 1(c) in the main manuscript.

Let us consider the case where the input $I(t)$ to the model described in Eqn.~\ref{eqn:1} is a white Gaussian noise with a constant mean value $\langle I \rangle$ and a correlation function $\langle (I(t)-\langle I \rangle) (I(t')-\langle I \rangle) \rangle=2D_I\delta(t-t')$ (here, we denote the mean of a parameter $H$ as $\langle H \rangle$). In accordance with \cite{vilela2009input}, we  make the following variable changes:
\begin{equation*}
v=\frac{U-U_{rest}}{V_{th}-U_{rest}}, t\longrightarrow \frac{t}{\tau_{m}}.
\tag{a}
\end{equation*}
When the membrane potential is $U$, the input current through the resistance branch of the RC circuit model becomes $(U-U_{rest})/R$. We denote the opposite of this input current as $I_{model} (U)=-(U-U_{rest})/R$. 
Taking the variable changes from (a) into account and differentiating $v$ with respect to time, 
we get:
\begin{equation*}
\frac{dv}{dt}=\dot v = \frac{\tau_m}{V_{th}-U_{rest}}\frac{dU}{dt}. 
\tag{b}
\end{equation*}
In addition, using the scaling property of the delta function \cite{proakis2001digital}, the correlation function of $I(t)$ becomes  $\frac{2D_I}{\tau_m}\delta(t-t')$ (since $\delta(\tau_mt)=\frac{1}{\tau_m} \delta(t)$).
Accordingly, we denote input $I(t)$ as follows:
\begin{equation*}
I(t)=\langle I \rangle+\sqrt{\frac{2D_I}{\tau_m}}\xi(t), 
\tag{c}
\end{equation*}
where $\xi(t)$ is a zero-mean white Gaussian noise with $\langle \xi(t) \xi(t') \rangle=\delta(t-t')$. Using the relations from (b) and (c) and dividing both sides by $(V_{th}-U_{rest})$ in Eqn.~\ref{eqn:1}, we obtain the following equation:
\begin{equation}\label{eqn:2s}
\dot v= -\frac{(U-U_{rest})}{V_{th}-U_{rest}} + \frac{R}{V_{th}-U_{rest}} \langle I \rangle  + \frac{R}{V_{th}-U_{rest}} \sqrt{\frac{2D_I}{\tau_m}}\xi(t). 
\end{equation}
Considering $v=\frac{U-U_{rest}}{V_{th}-U_{rest}},$ we acquire $(V_{th}-U_{rest})v+U_{rest}=U$. Therefore, we can get:
\begin{equation*}
I_{model} ((V_{th}-U_{rest})v+U_{rest})=I_{model} (U)=-\frac{U-U_{rest}}{R}
\tag{3}
\end{equation*}
Based on Eqn. 3, the first term on the right-hand side in Eqn.~\ref{eqn:2s} can be written as:
\begin{align*}
\frac{R}{V_{th}-U_{rest}}\frac{-(U-U_{rest})}{R}&=\frac{R}{V_{th}-U_{rest}}[I_{model} ((V_{th}-U_{rest})v+U_{rest})]\\
&=\frac{R}{V_{th}-U_{rest}}[I_{model} ((V_{th}-U_{rest})v+U_{rest})-I_{model}(U_{rest})]\\
&+\frac{R}{V_{th}-U_{rest}}[I_{model}(U_{rest})]. 
\end{align*}
Next, by merging the time-invariant term, $\frac{R}{V_{th}-U_{rest}}[I_{model}(U_{rest})]$ with the $\frac{R}{V_{th}-U_{rest}} \langle I \rangle$ term on the right-hand side in Eqn.~\ref{eqn:2s}, we can define the $f_{model} (v)$, $\mu$ and $D$ as presented in \cite{vilela2009input}:
\begin{equation*}
f_{model} (v)=\frac{R}{V_{th}-U_{rest}}[I_{model} ((V_{th}-U_{rest})v+U_{rest})-I_{model}(U_{rest})],
\tag{d}
\end{equation*}
\begin{equation*}
\mu=\frac{R}{V_{th}-U_{rest}}[\langle I \rangle  +I_{model}(U_{rest})],
\tag{e}
\end{equation*}
and 
\begin{equation*}
D=\frac{D_IR^2}{\tau_m (V_{th}-U_{rest})^2}. 
\tag{f}
\end{equation*}
Here, $\mu$ and $D$ are input parameters that represent the mean and the intensity of the fluctuating input in our model, respectively. Using the definitions from (d), (e) and (f), Eqn.~\ref{eqn:2s} can be further written as follows:
\begin{equation*}\label{eqn:4}
\dot v= f_{model}(v) + \mu  +\sqrt{2D}\xi(t). 
\tag{4}
\end{equation*}
Now, $I_{model} (U_{rest})=-(U_{rest}-U_{rest})/R=0$. 
Therefore, from Eqn. (d),
$f_{model}(v)$ for the LIF model can be transformed as follows:

\begin{align*}
f_{LIF}=f_{model}(v)&=\frac{R}{V_{th}-U_{rest}}[I_{model} ((V_{th}-U_{rest})v+U_{rest})-I_{model}(U_{rest})]\\
&=\frac{R}{V_{th}-U_{rest}}[I_{model} ((V_{th}-U_{rest})v+U_{rest})];[\because I_{model} (U_{rest})=0]\\
&=\frac{R}{V_{th}-U_{rest}} \frac{-(U-U_{rest})}{R};[using~Eqn.~3]\\
&=\frac{-(U-U_{rest})}{V_{th}-U_{rest}}=-v;[from (a)],
\end{align*}

Therefore, Eqn.~\ref{eqn:4} becomes as follows:
\begin{equation*}\label{eqn:5}
\dot v= -v + \mu  +\sqrt{2D}\xi(t),
\tag{5}
\end{equation*}
which is the formalism also used in \cite{lindner2014low} and 
will be followed for the remaining discussions in this study. 

\subsection{Coherence function}

Our analysis is based on the parallel $RC$ circuit model of the LIF neuron \cite{epfl} as depicted by Eqn.~\ref{eqn:1}. Here, the membrane capacitance $C$ integrates the input currents over time and the resistance branch $R$ represents the leakage path of membrane potential. For IF neuron model, since there is no leak path, the $R$ branch is considered as an open circuit. Hence, in this case, $I_{model}=\frac{-(U-U_{rest})}{R}=0$ and the RC circuit model only contains the capacitor $C$ path. This implies that, for the IF model, $R$ becomes infinity, and correspondingly the membrane time constant $\tau_m$, which is equal to $RC$, also becomes infinity. On the other hand, for LIF neuron model, the $R$ branch plays a role as the leakage path of membrane potential. When the leakage current through the resistance path increases, the resistance value $R$ and the membrane time constant $\tau_m$ gradually decrease. Furthermore, the parameters $\mu$ and $D~(=\frac{D_IR^2}{\tau_m (V_{th}-U_{rest})^2}=\frac{D_IR}{C (V_{th}-U_{rest})^2})$ become proportional to $R$ according to Eqns. (e) and (f), respectively. Therefore, for the LIF neuron models, $D$ and $\mu$ gradually decrease with the increase in leak amount. 

The author in \cite{lindner2014low} considered $D=D_{bg}+D_{st}$ in Eqn.~5, where $D_{bg}$ is the background noise intensity, $D_{st}$ is the intensity of the stimulus (Gaussian white noise input) and $D$ is the total noise intensity. 
For our analysis, by assuming $D_{bg}=0$, we get $D=D_{st}$ (note, a similar consideration was made in \cite{lindner2014low} for the results and analysis). Now, let us consider the output spike train of the model described by Eqn.~\ref{eqn:5} is $x(t)=\sum \delta[t-t_{k}]$, where $t_{k}$ is the $k^{th}$ instant of spike timing, when the input stimulus (s) is Gaussian white noise input. We quantify the information transmission of the spiking model by means of the spectral coherence function. To that end, the Fourier transform of $x(t)$ in a time window $[0,T]$ becomes as follows: $\tilde{x}_T(\omega)=\int_{0}^{T} x(t)e^{j\omega t} dt$. The cross-spectrum of output spike train (x) and input stimulus (s) is given as \cite{proakis2001digital}: $S_{x,s}(\omega)=\lim_{T\to\infty} \frac{\langle \tilde{x}(\omega)\tilde{s}^*(\omega) \rangle}{T},$ and the spike train power spectrum is defined as: $S_{x,x}(\omega)=\lim_{T\to\infty} \frac{\langle \tilde{x}(\omega)\tilde{x}^*(\omega) \rangle}{T}.$ The coherence function is formally defined as the squared correlation coefficient between the input and output as follows:
\begin{equation*}
C_{x,s}(\omega)=\frac{|S_{x,s}(\omega)|^2}{S_{x,x}(\omega)S_{s,s}(\omega)}.
\tag{6}
\end{equation*}
The coherence function $C_{x,s}(\omega) $ generates an output number between 0 and 1 at each measurement frequency. The amount of information transmission at each frequency is proportional to the coherence at that particular frequency, with 1 and 0 denoting full and null transmission, respectively. For a system acting as a low-pass filter, the coherence output under white-noise stimulation decreases in the high frequency domain.

Next, we analyze the low-pass filtering effect of the LIF neuron model as described by Eqn.~5. The analytical expression for $S_{x,s}(\omega)$ is given as follows \cite{lindner2014low, lindner2001transmission, lindner2004effects}:
\begin{equation*}
S_{x,s}(\omega)= \frac{2D_{st}}{\sqrt{D}}\frac{r_0i\omega}{i\omega-1}\frac{\mathscrsfs{D}_{i\omega-1} \left(\frac{\mu-V_{th}}{\sqrt{D}}  \right)-e^\Delta \mathscrsfs{D}_{i\omega-1} \left(\frac{\mu-U_{rest}}{\sqrt{D}}  \right)}{\mathscrsfs{D}_{i\omega} \left(\frac{\mu-V_{th}}{\sqrt{D}}  \right)-e^\Delta e^{i\omega \tau_r} \mathscrsfs{D}_{i\omega} \left(\frac{\mu-U_{rest}}{\sqrt{D}}  \right)},
\tag{7}
\end{equation*}
where $\Delta= \frac{U_{rest}^2-V_{th}^2+2\mu (V_{th}-U_{rest})}{4D}$, $\tau_r$ is the refractory period and $\mathscrsfs{D}(x)$ is the parabolic cylinder function. In our case, we follow the same assumptions as in \cite{lindner2014low} where $U_{rest}=0$, $\tau_r=0$ and $V_{th}=1$. The firing rate $r_0$ is given by calculating the following \cite{lindner2014low}:
\begin{equation*}
r_0= \left [ \tau_r + \sqrt{\pi}  \int_{\frac{\mu-V_{th}}{\sqrt{2D}}}^{\frac{\mu-U_{rest}}{\sqrt{2D}}} dz e^{z^2} erfc(z)   \right]^{-1}.
\end{equation*}
The power spectrum of the output spike train is given by \cite{lindner2002maximizing}, calculated as follows:
\begin{equation*}
S_{x,x}(\omega)= r_0\frac{\left |\mathscrsfs{D}_{i\omega} \left(\frac{\mu-V_{th}}{\sqrt{D}}  \right)\right|^2-e^{2\Delta} \left |\mathscrsfs{D}_{i\omega} \left(\frac{\mu-U_{rest}}{\sqrt{D}}  \right) \right |^2} { \left |\mathscrsfs{D}_{i\omega} \left(\frac{\mu-V_{th}}{\sqrt{D}}  \right)-e^\Delta e^{i\omega \tau_r} \mathscrsfs{D}_{i\omega} \left(\frac{\mu-U_{rest}}{\sqrt{D}}  \right)\right |^2},
\tag{8}
\end{equation*}
and the noise input spectrum becomes \cite{lindner2014low}:
\begin{equation*}
S_{s,s}(\omega)=2D_{st}. 
\tag{9}
\end{equation*}
Taking the magnitude square of the quantity in Eqn.~7, we derive the following,
\begin{equation*}
|S_{x,s}(\omega)|^2= \frac{4D_{st}^2}{{D}}\frac{r_0^2\omega^2}{1+\omega^2}\frac{|\mathscrsfs{D}_{i\omega-1} \left(\frac{\mu-V_{th}}{\sqrt{D}}  \right)-e^\Delta \mathscrsfs{D}_{i\omega-1} \left(\frac{\mu-U_{rest}}{\sqrt{D}}  \right)|^2}{|\mathscrsfs{D}_{i\omega} \left(\frac{\mu-V_{th}}{\sqrt{D}}  \right)-e^\Delta e^{i\omega \tau_r} \mathscrsfs{D}_{i\omega} \left(\frac{\mu-U_{rest}}{\sqrt{D}}  \right)|^2}.
\tag{10}
\end{equation*}

Finally, plugging the values of $S_{x,x}(\omega),~S_{s,s}(\omega)$ and $|S_{x,s}(\omega)|^2$ into Eqn.~6, we obtain the resultant coherence function as follows:

\begin{equation*}
\small{
C_{x,s}(\omega)=\frac{2D_{st}}{{D}}\frac{r_0\omega ^2}{1+\omega^2}\frac{\left |\mathscrsfs{D}_{i\omega-1} \left(\frac{\mu-V_{th}}{\sqrt{D}}  \right)-e^\Delta \mathscrsfs{D}_{i\omega-1} \left(\frac{\mu-U_{rest}}{\sqrt{D}}  \right)\right |^2}{\left |\mathscrsfs{D}_{i\omega} \left(\frac{\mu-V_{th}}{\sqrt{D}}  \right)\right|^2-e^{2\Delta} \left |\mathscrsfs{D}_{i\omega} \left(\frac{\mu-U_{rest}}{\sqrt{D}}  \right) \right |^2}}.
\tag{11}
\end{equation*}

\newpage
\section{Pseudo code for Spike-based Backpropagation Learning}
Here, we present the pseudo-code for the spike-based backpropagation learning used for training multi-layer SNNs in this work.

\begin{algorithm}[h]
   \caption{Procedure of spike-based backpropagation learning for an iteration.}
   \label{alg:surrogate}
\begin{algorithmic}
   \STATE {\bfseries Input:} pixel-based inputs ($inputs$), total number of time steps ($\#timesteps$), number of layers ($L$), weights ($W$), membrane potential ($U$), membrane time constant ($\tau_m$), firing threshold ($V_{th}$)
   \STATE {\bfseries Initialize:} $U_l[t] = 0,~\forall l = 1,...,L$
   \STATE //~{\bfseries Forward Phase}
   \FOR{$t \leftarrow 1$ {\bfseries to} $\#timesteps$}
        \STATE //~generate Poisson spike-inputs of a mini-batch data  
        \STATE $O_1[t] = Poisson(inputs);$
        \FOR{$l \leftarrow 2$ {\bfseries to} $L-1$}         
            \STATE //~membrane potential integrates weighted sum of spike-inputs
            \STATE $U_l[t]~= U_l[t-1] + W_{l} O_{l-1}[t]$
        \IF {$U_l[t] > V_{th}$}
        \STATE //~if membrane potential exceeds $V_{th}$, a neuron fires a spike
            \STATE $O_{l}[t] = 1,~U_l[t] = 0$
        \ELSE
        \STATE //~else, membrane potential decays exponentially
            \STATE $O_{l}[t] = 0,~U_l[t] = e^{-\frac{1}{\tau_m}} * U_l[t]$
        \ENDIF
        \ENDFOR    
        \STATE //~final layer neuron does not fire
        \STATE $U_{L}[t]~= e^{-\frac{1}{\tau_m}} * U_{L}[t-1] + W_{L} O_{L-1}[t]$
   \ENDFOR
   \STATE //~{\bfseries Backward Phase}
   \FOR{$t \leftarrow \#timesteps$ {\bfseries to} $1$}
       \FOR{$l \leftarrow L-1$ {\bfseries to} $1$}    
       \STATE //~evaluate partial derivatives of loss with respect to weight by unrolling the network over time
       \STATE $\triangle W_l[t] = \frac{\partial Loss}{\partial O_l[t]}\frac{\partial O_l[t]}{\partial U_l[t]}\frac{\partial U_l[t]}{\partial W_l[t]}$
       \ENDFOR
   \ENDFOR
   \end{algorithmic}
\end{algorithm}

\end{document}